%% file: main.tex
\documentclass[letterpaper]{article} 
\usepackage[preprint]{aaai2027}  

\usepackage[hyphens]{url}  
\usepackage{graphicx} 
\usepackage{natbib}  
\usepackage{caption} 
\usepackage{algorithm}
\usepackage{algorithmic}

\usepackage{newfloat}
\usepackage{listings}
\DeclareCaptionStyle{ruled}{labelfont=normalfont,labelsep=colon,strut=off} 
\floatstyle{ruled}
\newfloat{listing}{tb}{lst}{}
\floatname{listing}{Listing}

\usepackage{booktabs}

\usepackage{multirow}
\usepackage{amsmath}
\usepackage{amssymb}
\usepackage{array}
\usepackage{tabularx}

\title{\methodname: A Unified Agentic System for Intelligent Document Processing}
\author{
    Siqi Xiang\textsuperscript{\rm 1},
    Zhipeng Xu\textsuperscript{\rm 2},
    Yufei Liu\textsuperscript{\rm 1},
    Junhao Ji\textsuperscript{\rm 2},
    Qing Liu\textsuperscript{\rm 2},
    Zulong Chen\textsuperscript{\rm 2},
    Zhibo Yang\textsuperscript{\rm 3},
    Chunyan Miao\textsuperscript{\rm 1},
    Shijian Lu\textsuperscript{\rm 1}\corresponding
}
\affiliations{

    \textsuperscript{\rm 1}Nanyang Technological University\\
    \textsuperscript{\rm 2}Alibaba Group\\
    \textsuperscript{\rm 3}Qwen Team, Alibaba Group\\
}

\newcommand{\methodname}{DocClaw}

\begin{document}

\maketitle

\thispagestyle{plain}

\begin{abstract}
\input{sections/abstract}
\end{abstract}


\input{sections/introduction}

\input{sections/related_work}

\input{sections/framework}

\input{sections/experiment}

\input{sections/conclusion}

\bibliography{aaai2027}


\end{document}

%% file: sections/abstract.tex
Intelligent document processing (IDP) encompasses a broad range of tasks, including optical character recognition (OCR), document question answering (DocQA), and key information extraction (KIE). Despite their distinct objectives, these tasks share a common need to perceive document content, acquire task-relevant information, and progressively refine intermediate results. However, they are typically formulated as separate prediction problems and addressed by task-specific models or processing pipelines.
We introduce \methodname{}, a unified agentic system that formulates diverse intelligent document processing tasks as a shared process of interaction between an agent and a document. Given a document and a task-specific query, \methodname{} follows an appropriate document skill to iteratively identify the information required, invoke relevant tools, and integrate the resulting observations into the desired output. Throughout this process, a structured document state organizes reusable document knowledge and task-specific interaction context, allowing the agent to accumulate, revisit, and progressively refine information as the interaction proceeds. Under this formulation, task-specific requirements are captured by the agent's interpretation of the query objective and the corresponding document skill, while the underlying interaction loop, tool space, and document state are shared across tasks.
Extensive experiments across multiple intelligent document processing benchmarks demonstrate that \methodname{} effectively handles diverse tasks within a single agentic framework and achieves competitive performance compared with both general-purpose VLMs and task-specific methods.

%% file: sections/introduction.tex
\section{Introduction}
\label{sec:introduction}

\begin{figure}[t]
\centering
\includegraphics[width=0.98\columnwidth]{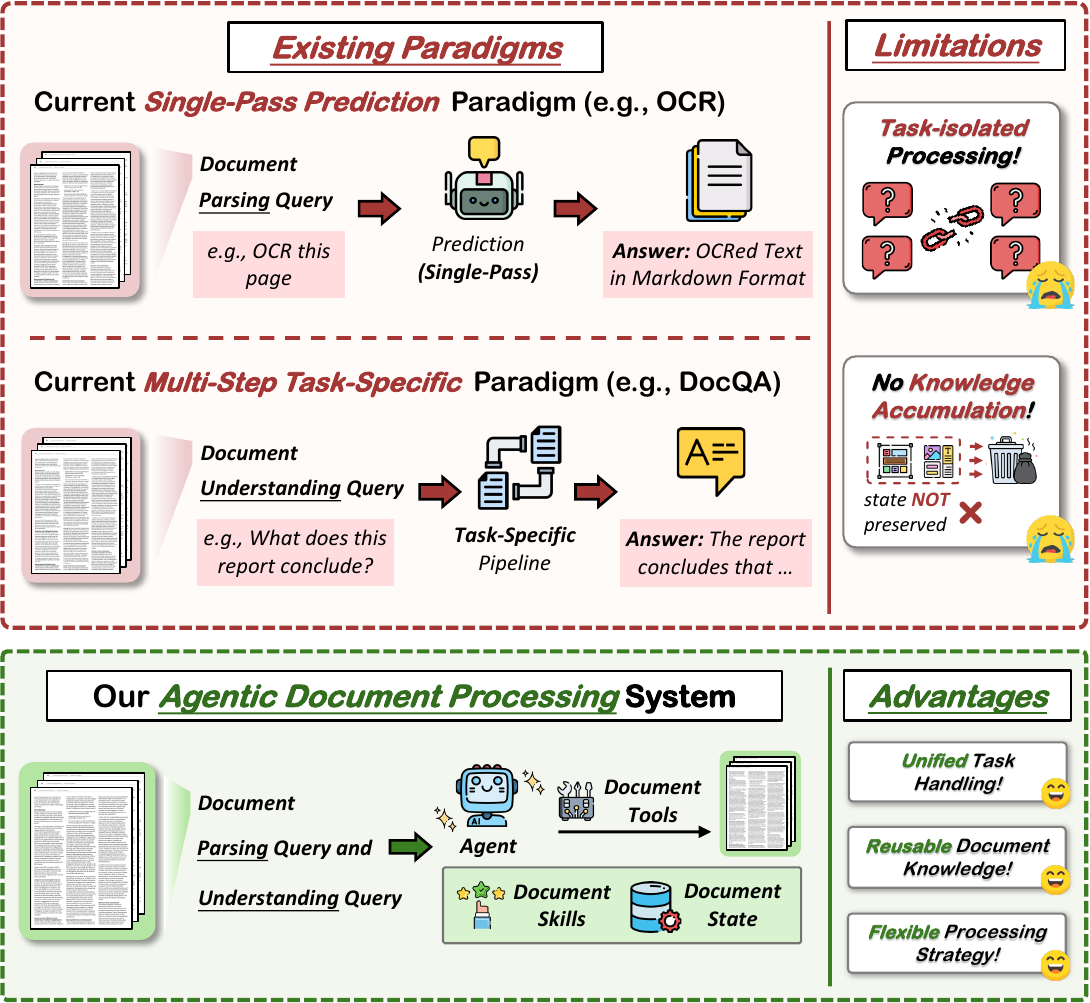}

\caption{Unlike existing single-pass and multi-step paradigms, the proposed \methodname{} formulates intelligent document processing as interactions between an agent and a document, leveraging iterative tool invocation and knowledge accumulation across queries.}
\label{fig:method_comparison}
\end{figure}

Intelligent document processing (IDP) aims to extract and reason over information jointly conveyed through textual content, spatial layout, and visual elements~\cite{appalaraju2021docformer,li2021structext,tang2023udop}. It encompasses a broad spectrum of tasks, ranging from fundamental document perception tasks such as optical character recognition (OCR) to higher-level document understanding tasks such as document question answering (DocQA) and key information extraction (KIE)~\cite{li2021trocr,layoutlm,ji2026unikie}.
Existing IDP methods are predominantly developed around individual task objectives, resulting in separate processing chains for different document processing tasks. As illustrated in Figure~\ref{fig:method_comparison}, these methods typically follow one of two processing paradigms. Some methods adopt single-pass prediction, directly producing task-specific outputs without intermediate processing steps~\cite{ye2023ureader,xu2026enhancing,Qianfan-OCR}.
Others rely on multi-step task-specific pipelines, which decompose document processing into predefined sequences of specialized operations, progressively transforming intermediate results into the required outputs. By separating layout analysis from region-level recognition, these pipelines have achieved promising performance in document parsing~\cite{Dolphin,GLM-OCR,MonkeyOCR,PaddleOCR-VL-1.6,MinerU2.5-Pro}.

Despite these advances, most existing work treats intelligent document processing as a collection of isolated tasks, leading to two major limitations.
First, there remains an absence of a general framework for handling heterogeneous document processing tasks within a unified workflow.
Single-pass models rely on rigid, task-specific mappings, while multi-step pipelines remain constrained by predefined task-specific procedures.
As a result, neither paradigm can flexibly adapt its processing pattern across diverse intelligent document processing tasks.
Second, processing each task independently hinders document knowledge from being accumulated, reused, and refined across queries.
Document processing often involves interdependent workloads, where intermediate results can naturally serve as reusable knowledge for subsequent operations.
However, isolated task processing provides no unified mechanism for sharing such knowledge.

To address the above limitations, we introduce \methodname{}\footnote{Code is available at \url{https://github.com/sxiangag/DocClaw}.}, a unified agentic document processing system that reformulates document processing as iterative interactions with documents.
Through an iterative interaction loop, \methodname{} actively explores document content, selects and invokes appropriate document processing tools, updates the document state with newly acquired knowledge, and progressively refines intermediate results into task-specific outputs.
To support this iterative process, \methodname{} maintains a structured document state that serves as the agent's memory. The document state comprises document memory, which preserves document knowledge for reuse across queries with different objectives, and task memory, which tracks task-specific context throughout the current interaction.
To guide the agent's processing strategy across different query objectives, we introduce document skills, a set of task-oriented instructions that define document interaction strategies and tool-use patterns. For each query, \methodname{} selects the skill corresponding to its objective to guide subsequent planning and tool invocation.

We evaluate \methodname{} across three representative intelligent document processing tasks: optical character recognition (OCR), document question answering (DocQA), and key information extraction (KIE).
Experiments show that \methodname{} achieves competitive performance against both general-purpose vision-language models and task-specific methods, demonstrating its effectiveness in supporting heterogeneous document processing tasks within a unified agentic framework.
Further ablation studies demonstrate that the document state enables effective accumulation and reuse of document knowledge, whereas document skills provide task-oriented guidance that steers the agent toward task-specific processing strategies, allowing intermediate results to be progressively improved through dedicated refinement operations tailored to different query objectives.
Our main contributions are summarized as follows:
\begin{itemize}

\item We introduce \methodname{}, a unified agentic framework that formulates intelligent document processing as iterative interactions between an agent and a document, allowing different query objectives to be addressed within a shared interaction loop and document processing tool space.

\item We propose a structured document state that organizes persistent document knowledge and task-specific context, supporting progressive knowledge accumulation, reuse, and refinement across interactions and queries.

\item We introduce document skills, a set of task-oriented instructions that guide the agent toward appropriate processing strategies and tool-use patterns for different query objectives, together with dedicated refinement operations that progressively improve intermediate results.

\end{itemize}

%% file: sections/related_work.tex
\section{Related Work}
\label{sec:related_works}

\paragraph{Intelligent Document Processing}
Intelligent document processing aims to derive structured or semantic outputs from the textual and visual information embedded in a document.
Existing intelligent document processing methods can be broadly categorized into single-pass prediction and multi-step task-specific pipelines.
Single-pass prediction utilizes specialized models for individual tasks, such as optical character recognition (OCR) and document layout analysis (DLA).
These models directly predict task-specific outputs from document images, with their architectures and training objectives designed for corresponding task formulations~\cite{Qianfan-OCR,HunyuanOCR}.
While effective for their target tasks, they are generally difficult to adapt to heterogeneous document processing objectives.
Multi-step task-specific pipelines decompose document processing into a sequence of specialized operations.
For example, PaddleOCR-VL~\cite{PaddleOCR-VL-1.6} first performs layout analysis to identify and order document regions, then recognizes individual regions and assembles their outputs into a complete document transcription.
Similar multi-step pipelines have also been explored for document understanding, where systems retrieve document content, inspect candidate pages, and generate task-specific outputs~\cite{DocAgent,mdocagent,docdancer}.
However, their operations are typically organized into predefined chains designed for specific tasks.
Reusable capabilities such as layout analysis, text recognition, and information retrieval remain embedded in separate task-specific pipelines. In contrast, \methodname{} organizes these capabilities within a unified agentic framework that dynamically plans and invokes document processing tools according to the query objective and document state.

\paragraph{Agentic and Tool-Augmented Reasoning.}

Agentic and tool-augmented reasoning has emerged as an effective paradigm for extending language models beyond text generation.
Retrieval-augmented methods ground model outputs in external evidence, while tool-augmented approaches enable models to invoke external tools and incorporate the resulting observations into subsequent reasoning~\citep{RAG,react,Toolformer}.
Multi-agent frameworks further extend this interaction paradigm by coordinating agents with different roles and capabilities to collaboratively solve complex tasks~\citep{CAMEL,AutoGen,ChatEval}.
Despite these advances, most existing agentic systems are designed for general-purpose reasoning and treat external tools as independent resources invoked to solve a particular task. When applied to document processing, they lack a unified mechanism for coordinating heterogeneous document tools and maintaining reusable document knowledge across interactions.
Consequently, document knowledge acquired during parsing, retrieval, and reasoning is difficult to accumulate, refine, and reuse over time.
To address this limitation, \methodname{} introduces a unified agentic framework that coordinates document tools through an agentic interaction loop and maintains a structured document state to accumulate, refine, and reuse document knowledge across interactions.

%% file: sections/framework.tex
\section{Methodology}
\label{sec:method}

\begin{figure*}[t]
\centering
\includegraphics[width=0.98\textwidth]{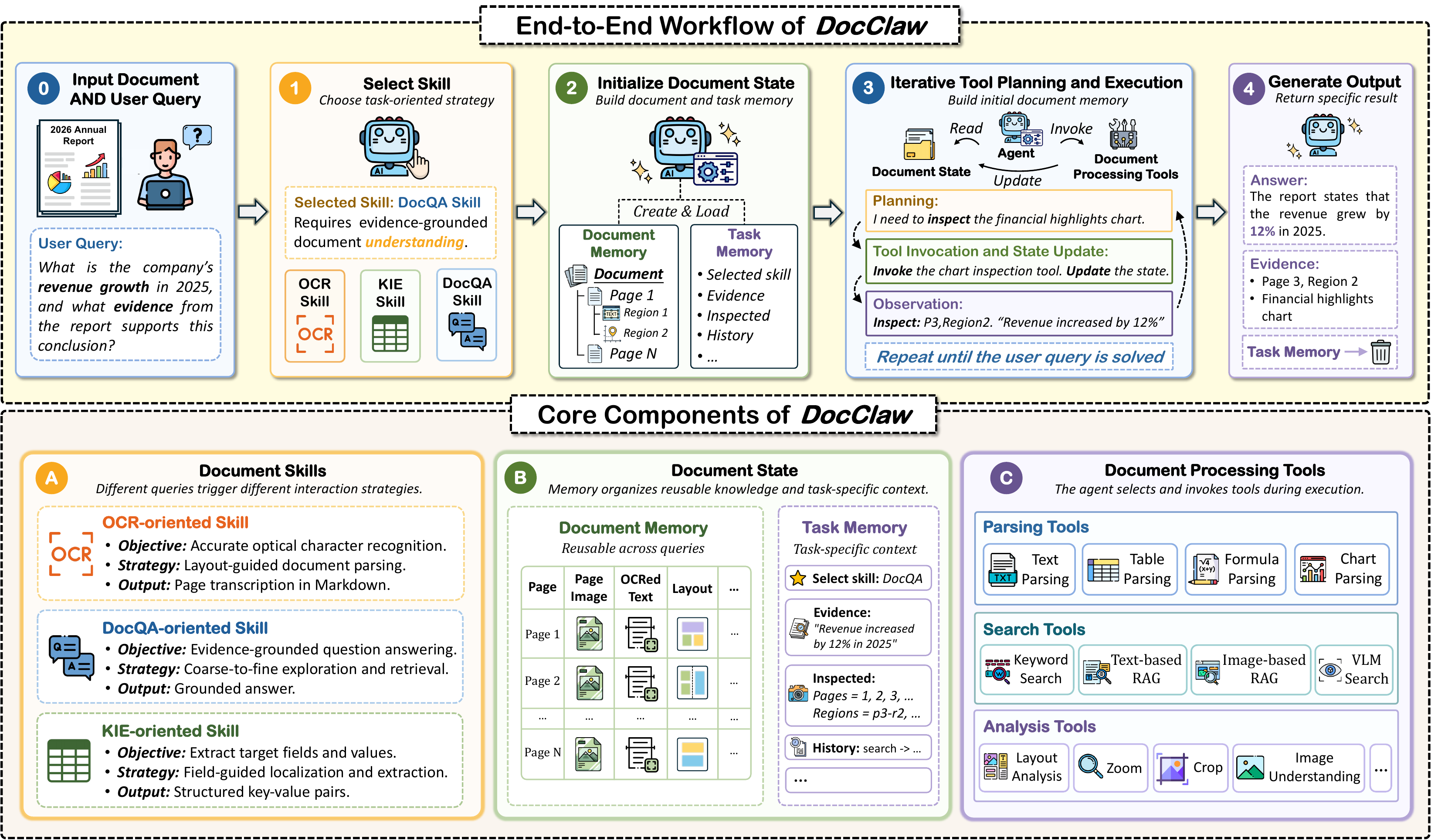}
\caption{Overview of the proposed \methodname{}. Given an input document and a user query, \methodname{} first selects a relevant task-oriented document skill and initializes the document state for the current query. It then iteratively plans and executes document processing tools, updating the document state until the final task-specific output is generated.}
\label{fig:docclaw_overview}
\end{figure*}

This section details our proposed framework, \methodname{}.
We first formulate intelligent document processing as iterative interactions between an agent and a document.
We then present the document state, a structured memory that organizes document knowledge and task-specific interaction context throughout the agentic interaction process.
Finally, we introduce document skills, a set of task-oriented instructions that guide the agent's interaction strategies and tool-use patterns under different query objectives.

\subsection{Agentic Document Processing}
\label{sec:method_agentic_interaction}

Formally, given an input document $D$ and a query $q$, the goal of intelligent document processing is to produce a task-specific output $y$ in response to $q$.
Depending on the query objective, $y$ may take different forms, such as recognized document content for OCR, extracted field-value pairs for KIE, or answers to questions for DocQA.
\methodname{} provides a unified formulation for these heterogeneous objectives by treating document processing as iterative interactions between an agent and the document, as illustrated in Figure~\ref{fig:docclaw_overview}.

For each user query $q$, \methodname{} first selects a document skill $k \in \mathcal{K}$ from a predefined skill set according to the query objective.
The selected skill specifies the task-oriented interaction strategy and tool-use patterns that guide subsequent processing.
The document state is then initialized as $s_0=(m_0,e_0)$, consisting of document memory $m_0$ and task memory $e_0$.
For a previously unseen document, the document memory is initialized from its page images $\{p_i\}_{i=1}^{N}$, whereas for a document that has already been processed, the retained document memory serves as the starting state for the new query.
In contrast, the task memory is initialized independently for each query and starts empty, i.e., $e_0=\varnothing$.

After initialization, an LLM-based agent interacts with the document state using document processing tools through a ReAct-style~\cite{react} agentic loop.
At each interaction step $t$, the agent first reads a condensed representation $\phi(s_t)$ of the current document state, which summarizes what document information has already been acquired and what task-specific interaction progress has been recorded.
Conditioned on $\phi(s_t)$, the user query $q$, and the selected document skill $k$, the planner determines the next document processing action:
\begin{equation}
\begin{gathered}
a_t = (\tau_t,c_t)
=\pi_{\theta}\!\left(q,k,\phi(s_t)\right), \\[3pt]
\tau_t\in\mathcal{T},
\qquad
c_t \in \mathcal{C}(\tau_t),
\end{gathered}
\label{eq:tool_planning}
\end{equation}
where $\tau_t$ denotes the selected document processing tool, $c_t$ specifies its invocation arguments, and $\mathcal{C}(\tau_t)$ denotes the valid argument space of tool $\tau_t$.
This allows the agent to dynamically determine the document operation to perform and its corresponding invocation arguments based on the current document state.

The planned action is then executed on the document and produces an execution result $r_t$.
Rather than being used as a transient observation, the execution result is incorporated into the document state to support subsequent interactions and preserve acquired document knowledge for future queries.
We formulate the tool execution and state update as:
\begin{equation}
\begin{aligned}
r_t
    &= \mathcal{F}_{\tau_t}
    \left(D,c_t;s_t\right), \\[2pt]
s_{t+1}
    &= \mathcal{U}(s_t,a_t,r_t) \\
    &=
    \Big(
        \mathcal{U}_m(m_t,a_t,r_t),
        \mathcal{U}_e(e_t,a_t,r_t)
    \Big),
\end{aligned}
\label{eq:tool_execution_update}
\end{equation}

where $\mathcal{F}_{\tau_t}$ denotes the execution of the selected tool $\tau_t$ with invocation configuration $c_t$, and $\mathcal{U}$ denotes the corresponding state update.
The update functions $\mathcal{U}_m$ and $\mathcal{U}_e$ integrate reusable document knowledge and task-specific interaction context into document memory and task memory, respectively.
Specifically, document knowledge acquired through tool execution, such as recognized content and layout structures, is accumulated in document memory, while task-specific context is recorded in task memory.
Through this process, successive tool calls progressively enrich the document state, allowing information accumulated from earlier steps to support subsequent interactions and future queries.

After the state update, the planner reads $\phi(s_{t+1})$ of the updated document state to determine the next document-processing action with reference to the user query $q$ and the selected skill $k$.
The interaction continues until the agent determines that sufficient information has been acquired to satisfy the query and selects an output action $a_{\mathrm{out}}$.
For the $j$-th query $q^{(j)}$ issued on the document $D$, the final output and the transition to the next query are formulated as:
\begin{equation}
\begin{gathered}
a_{T_j}^{(j)} = a_{\mathrm{out}}, \\[3pt]
y^{(j)}
= \mathcal{F}_{\mathrm{out}}
\left(
    q^{(j)},
    k^{(j)},
    s_{T_j}^{(j)}
\right), \\[3pt]
m_0^{(j+1)} = m_{T_j}^{(j)},
\qquad
e_0^{(j+1)} = \varnothing,
\end{gathered}
\label{eq:cross_query_transition}
\end{equation}
where $T_j$ denotes the final interaction step for query $q^{(j)}$.
After the output is generated, the task memory for the completed query is discarded, while the document memory is retained.
This allows subsequent queries to reuse previously acquired document knowledge, reducing redundant tool execution while avoiding irrelevant task-specific context.

Through this iterative process, heterogeneous intelligent document processing queries share a unified agentic interaction loop, while task-oriented skills guide tool usage and persistent state updates enable document knowledge to be progressively accumulated, refined, and reused across queries.

\subsection{Document State}
\label{sec:method_document_state}

During iterative agentic interactions, tool execution produces both reusable document knowledge and task-specific context.
To organize these two types of information, \methodname{} maintains a structured document state:
\begin{equation}
    s_t = \left(m_t,e_t\right),
\end{equation}
where $m_t$ denotes persistent document memory and $e_t$ denotes query-level task memory.
This separation allows tool-acquired document knowledge to be continuously reused and refined while preventing task-specific context from carrying over to subsequent interactions.

\paragraph{Document Memory.}
The document memory $m_t$ organizes document knowledge acquired throughout agentic interactions. To preserve the intrinsic structure of documents, \methodname{} organizes this knowledge hierarchically at the document, page, and region levels.
The document level maintains global information about the document and its constituent pages.
Each page stores its raw page image together with page-level information acquired during processing, such as OCRed text, layout structure, and reading order.
When regions are identified, they are associated with their corresponding pages and maintain region-level information, including spatial coordinates, region type, recognized text, and other tool-produced attributes.

This hierarchical organization provides a shared structure through which different document processing tools can progressively add and refine document knowledge.
For example, layout analysis identifies page regions and their structural properties, while recognition tools attach recognized content to the corresponding pages and regions.
Document knowledge acquired by different tools is therefore consolidated around shared document entities and remains reusable across queries.
During planning, the agent accesses the document state through $\phi(s_t)$, which retains only the information necessary for planning and tool selection. Therefore, contents such as page images and recognized text remain accessible to document processing tools while being omitted from the planner context, reducing planning context overhead.

\paragraph{Task Memory.}

The task memory $e_t$ maintains the query-specific context throughout agentic interactions.
Specifically, it records the selected document skill and progressively accumulates information relevant to resolving the query, such as retrieved evidence, inspected pages or regions, intermediate results, and interaction history.
As tool execution proceeds, these entries are continuously updated to reflect the evolving context of the current query.

As task memory contains detailed information relevant to solving the ongoing query, it provides the context required for subsequent tool planning.
Therefore, the task memory $e_t$ is fully exposed to the planner at each interaction step, allowing the agent to directly access accumulated evidence, inspected targets, and interaction history when determining the next action.
Once the final output is produced, the task memory is discarded. This prevents query-specific context from carrying over to subsequent queries.

\subsection{Document Skills}
\label{sec:method_document_skills}

While the agentic interaction formulates a unified execution workflow for intelligent document processing, queries with different objectives require distinct interaction strategies and tool-use patterns.
To accommodate these diverse objectives, \methodname{} lets the agent select a document skill from a predefined document skill set based on its interpretation of the query objective.
Each document skill provides task-specific guidance for subsequent planning and tool selection.

\paragraph{OCR-oriented Skill.}
The OCR-oriented skill is designed for accurate document transcription.
Similar to pipeline-based methods~\cite{PaddleOCR-VL-1.6,MinerU2.5-Pro}, it first guides layout analysis to identify document regions and routes different region types to specific parsing tools, including text recognition, formula parsing, and table parsing.
To further improve recognition quality, the skill encourages visual enhancement of low-quality regions and guides the agent to apply zoom, crop, and rotate operations before re-recognition.
Together, these instructions allow the agent to adapt its processing strategy to the content and quality of regions through specialized parsing and visual enhancement.

\paragraph{DocQA-oriented Skills.}
DocQA-oriented skills target reliable question answering through multimodal search and evidence extraction. Since distinct document-based questions may require either localized evidence inspection or broader document-level coverage, we divide the question-answering policy into two complementary skills, the DocQA inspection skill and the DocQA enumeration skill. The inspection skill adopts a coarse-to-fine searching strategy that progressively narrows candidate pages, inspects relevant content, and extracts supporting evidence before answer generation. In contrast, the enumeration skill prioritizes broad document-level coverage for questions that require document-wide reasoning. The skill encourages the agent to maintain a sufficiently broad candidate page scope, thereby reducing the risk of insufficient page coverage in document-wide tasks such as occurrence counting and item grouping.

\paragraph{KIE-oriented Skill.}
The KIE-oriented skill targets reliable extraction of requested fields from document pages. The skill first guides direct field extraction from the page image. To mitigate errors from purely visual extraction, it then invokes OCR to obtain textual content and performs a second extraction from the recognized text. For fields with inconsistent predictions, the skill further specifies conflict resolution using the original page image as a reference to select the more reliable value. Overall, the skill combines visual extraction, OCR-based extraction, and image-grounded verification to improve the accuracy of structured field-value extraction.

%% file: sections/experiment.tex
\section{Experiments}
\label{sec:experiments}

\subsection{Experimental Setup}
\label{sec:experiments_setup}

\paragraph{Benchmarks and Metrics.}

We evaluate \methodname{} on three intelligent document processing tasks: optical character recognition (OCR), document question answering (DocQA), and key information extraction (KIE).

The OCR evaluation is conducted on OmniDocBench v1.6~\cite{OmniDocBench}, a document parsing benchmark containing 1,651 PDF pages.
Following the official evaluation protocol, we report the overall score and category-specific metrics that cover different document modules.
Specifically, Text Edit measures normalized edit distance for recognized text, Formula CDM evaluates formula recognition, and Table TEDS and Table TEDS-S evaluate table recognition with and without considering cell content, respectively.
Reading Order Edit measures the correctness of predicted reading sequences. The overall score aggregates performance across text, formula, and table modules.

For DocQA, we evaluate \methodname{} on MMLongBench-Doc~\cite{MMLongBenchDoc}, a document question answering benchmark consisting of 135 PDF documents with an average of 47.5 pages per document.
Adhering to the official evaluation protocol, raw responses are first normalized through LLM-based answer extraction using GPT-4.1~\cite{gpt4_1} and evaluated by rule-based scoring against the ground-truth answers.
Each question receives a score between 0 and 1 according to the match between the extracted prediction and the ground-truth answer, and the final accuracy is computed as the average score over all questions.

For KIE, we evaluate \methodname{} on the KIE subsets of OCRBench~\cite{ocrbench} and OCRBench v2~\cite{ocrbenchv2}. OCRBench contains 200 KIE samples formulated as single-field question answering, where predictions are evaluated by checking whether the ground-truth answer appears in the normalized output. OCRBench v2 contains 400 English and 400 Chinese KIE samples, both requiring structured key-value extraction. Following the official evaluation protocol, predictions are parsed into normalized field-value pairs and evaluated by field-level exact matching.

\begin{table*}[!t]
\centering
\small
\setlength{\tabcolsep}{2.8pt}
\renewcommand{\arraystretch}{1.08}

\begin{tabular}{@{}
>{\centering\arraybackslash}p{1.6cm}|
l|
>{\centering\arraybackslash}p{0.64cm}|
>{\centering\arraybackslash}p{1.24cm}
c c c c c
@{}}

\toprule

\multicolumn{9}{c}{
    \textbf{(a) Optical Character Recognition}
} \\
\midrule

\textbf{Type} &
\textbf{Method} &
\multicolumn{2}{c|}{\textbf{Overall}$\uparrow$} &
\textbf{Text}$^{\mathrm{Edit}}\downarrow$ &
\textbf{Formula}$^{\mathrm{CDM}}\uparrow$ &
\textbf{Table}$^{\mathrm{TEDS}}\uparrow$ &
\textbf{Table}$^{\mathrm{TEDS\text{-}S}}\uparrow$ &
\textbf{Order}$^{\mathrm{Edit}}\downarrow$ \\
\midrule

\multirow{3}{*}{\shortstack{\textbf{General}\\\textbf{VLMs}}}
& GPT-5.2
& \multicolumn{2}{c|}{86.59}
& 0.114 & 88.21 & 82.95 & 87.93 & 0.193 \\

& Qwen3-VL-235B
& \multicolumn{2}{c|}{89.78}
& 0.063 & 92.55 & 83.07 & 86.75 & 0.166 \\

& Gemini-3-Pro
& \multicolumn{2}{c|}{92.91}
& 0.064 & 95.99 & 89.15 & 92.96 & 0.165 \\

\midrule

\multirow{5}{*}{\shortstack{\textbf{Specialized}\\\textbf{VLMs}}}

& Youtu-Parsing
& \multicolumn{2}{c|}{93.74}
& 0.044 & 93.63 & 92.02 & 95.00 & \textbf{0.116} \\

& Qianfan-OCR & \multicolumn{2}{c|}{93.90} & 0.040 & 95.08 & 90.53 & 93.31 & 0.130 \\

& GLM-OCR & \multicolumn{2}{c|}{95.22} & 0.044 & 97.18 & 92.83 & 95.39 & 0.133 \\

& MinerU2.5-Pro
& \multicolumn{2}{c|}{95.75}
& 0.036 & 97.45 & 93.42 & 95.92 & \underline{0.120} \\

& PaddleOCR-VL-1.6
& \multicolumn{2}{c|}{96.34}
& \textbf{0.033} & 97.53
& \underline{94.76} & \underline{97.10} & 0.128 \\

\midrule

\multirow{3}{*}{\shortstack{\textbf{Document}\\\textbf{Agent}}}
& DocClaw w/ GPT-5.2
& \multicolumn{2}{c|}{\underline{96.40}}
& 0.035 & \underline{98.01}
& 94.71 & \underline{97.10} & 0.129 \\

& DocClaw w/ Gemini-2.5-Flash
& \multicolumn{2}{c|}{96.38}
& \underline{0.034} & 97.87
& \underline{94.76} & \textbf{97.13} & 0.129 \\

& DocClaw w/ Gemini-2.5-Pro
& \multicolumn{2}{c|}{\textbf{96.45}}
& \underline{0.034} & \textbf{98.05}
& \textbf{94.79} & \textbf{97.13} & 0.129 \\

\midrule
\multicolumn{9}{c}{
    \textbf{(b) Document Question Answering}
} \\
\midrule

\textbf{Type} &
\textbf{Method} &
\textbf{ALL} &
\textbf{TXT} &
\textbf{LAY} &
\textbf{CHA} &
\textbf{TAB} &
\textbf{FIG} &
\textbf{UNA} \\
\midrule

\multirow{3}{*}{\shortstack{\textbf{General}\\\textbf{VLMs}}}
& GPT-4o
& 42.8 & 46.3 & 46.0 & 45.3 & 50.0 & 44.1 & 20.2 \\

& Gemini-2.5-Flash
& 49.6 & 44.0 & 53.2 & 46.0 & 43.9 & 48.2 & 56.7 \\

& Gemini-2.5-Pro
& 58.1 & 52.1 & \underline{62.1} & 55.5 & 55.3 & 54.0 & 59.9 \\

\midrule

\multirow{6}{*}{\shortstack{\textbf{Specialized}\\\textbf{Pipelines}}}

& MDocAgent w/ GPT-4o
& 42.0 & -- & -- & -- & -- & -- & -- \\

& DocAgent w/ GPT-4o
& 51.8 & -- & -- & -- & -- & -- & -- \\

& DocDancer w/ Gemini-2.5-Pro
& 56.3 & -- & -- & -- & -- & -- & -- \\

& DocDancer w/ GPT-5.2
& 57.0 & -- & -- & -- & -- & -- & -- \\

& SimpleDoc w/ Gemini-2.5-Pro
& 56.6 & 48.4 & 54.8 & 55.7 & 56.1 & 52.5 & 59.7 \\

& SimpleDoc w/ Claude-4-Sonnet
& 58.6 & 52.1 & 53.3 & \underline{58.3} & \underline{62.4} & 46.9 & \underline{66.5} \\

\midrule

\multirow{3}{*}{\shortstack{\textbf{Document}\\\textbf{Agent}}}

& DocClaw w/ GPT-5.2
& 59.3 & \textbf{61.0} & 60.1 & 58.1 & 60.4 & \underline{55.8} & 58.7 \\

& DocClaw w/ Gemini-2.5-Flash
& \underline{61.0} & 55.2 & 57.2 & 54.9 & \underline{63.0} & 55.5 & \textbf{66.8} \\

& DocClaw w/ Gemini-2.5-Pro
& \textbf{61.5} & \underline{56.4} & \textbf{62.9} & \textbf{60.1} & \textbf{64.8} & \textbf{60.3} & 56.0 \\

\midrule
\multicolumn{9}{c}{
    \textbf{(c) Key Information Extraction}
} \\
\midrule

\textbf{Type} &
\textbf{Method} &
\textbf{ALL} &
\multicolumn{2}{c}{\textbf{OCRBench KIE}} &
\multicolumn{2}{c}{\textbf{OCRBench v2 KIE (EN)}} &
\multicolumn{2}{c}{\textbf{OCRBench v2 KIE (CN)}} \\
\midrule

\multirow{3}{*}{\shortstack{\textbf{General}\\\textbf{VLMs}}}

& GPT-5.2
& 69.8
& \multicolumn{2}{c}{88.0}
& \multicolumn{2}{c}{76.1}
& \multicolumn{2}{c}{45.4} \\

& Gemini-2.5-Flash
& 76.2
& \multicolumn{2}{c}{91.8}
& \multicolumn{2}{c}{83.3}
& \multicolumn{2}{c}{53.6} \\

& Gemini-2.5-Pro
& 77.0
& \multicolumn{2}{c}{\underline{93.5}}
& \multicolumn{2}{c}{83.3}
& \multicolumn{2}{c}{54.3} \\

\midrule

\multirow{2}{*}{\shortstack{\textbf{Specialized}\\\textbf{VLMs}}}

& Qianfan-OCR
& 84.0
& \multicolumn{2}{c}{88.0}
& \multicolumn{2}{c}{81.6}
& \multicolumn{2}{c}{\textbf{82.3}} \\

& GLM-OCR
& 39.7
& \multicolumn{2}{c}{60.0}
& \multicolumn{2}{c}{29.3}
& \multicolumn{2}{c}{29.1} \\

\midrule

\multirow{3}{*}{\shortstack{\textbf{Document}\\\textbf{Agent}}}

& DocClaw w/ GPT-5.2
& 81.0
& \multicolumn{2}{c}{92.5}
& \multicolumn{2}{c}{76.6}
& \multicolumn{2}{c}{73.7} \\

& DocClaw w/ Gemini-2.5-Flash
& \underline{84.8}
& \multicolumn{2}{c}{\underline{93.5}}
& \multicolumn{2}{c}{\underline{84.2}}
& \multicolumn{2}{c}{76.8} \\

& DocClaw w/ Gemini-2.5-Pro
& \textbf{86.9}
& \multicolumn{2}{c}{\textbf{94.5}}
& \multicolumn{2}{c}{\textbf{87.5}}
& \multicolumn{2}{c}{\underline{78.7}} \\

\bottomrule

\end{tabular}

\caption{
Main results across optical character recognition, document question answering, and key information extraction.
Panels (a) to (c) report results on OmniDocBench v1.6, MMLongBench-Doc, and the KIE subsets of OCRBench and OCRBench v2, respectively.
For document question answering, TXT, LAY, CHA, TAB, FIG, and UNA denote text, layout, chart, table, figure, and unanswerable questions.
For key information extraction, ALL averages the results across OCRBench KIE and OCRBench v2 subsets.
The best and second-best results for each metric are highlighted in bold and underlined, respectively.
}
\label{tab:main_results}
\end{table*}

\paragraph{Implementation Details.}

We instantiate \methodname{} with three multimodal foundation models as agent backbones: GPT-5.2, Gemini-2.5-Flash, and Gemini-2.5-Pro. For each backbone, the same model is used for the planner and model-based document processing tools unless otherwise specified.

The document processing toolset spans three categories: parsing, search, and analysis.
Parsing tools recognize text, tables, formulas, and charts, primarily using PaddleOCR-VL-1.6~\cite{PaddleOCR-VL-1.6}.
Search tools support keyword search, text-based retrieval, image-based retrieval, and VLM-based visual search using BM25~\cite{bm25}, ColBERT~\cite{ColBERT}, ColPali~\cite{ColPali}, and the backbone VLM, respectively.
Analysis tools support layout analysis with PP-DocLayout v3~\cite{pp-doclayout}, image understanding using the backbone VLM, and visual refinement operations such as zoom, crop, and rotate.
Other model-based tools perform OCR quality inspection, evidence extraction, answer generation, and candidate verification using the same backbone as the planner.
All experiments are conducted on a server equipped with an Intel Xeon Platinum 8369B CPU and two NVIDIA A100 80GB GPUs.

\subsection{Main Results}
\label{sec:experiments_main_results}

\paragraph{Optical Character Recognition.}
We report OCR results on OmniDocBench v1.6 in Table~\ref{tab:main_results}(a). \methodname{} consistently outperforms all baselines across all three backbone models.
Particularly, \methodname{} with Gemini-2.5-Pro obtains the best overall score of 96.45, compared with PaddleOCR-VL-1.6 at 96.34.
Notably, \methodname{} achieves the highest scores on Formula CDM, Table TEDS, and Table TEDS-S. These improvements are due to the targeted refinement of low-quality regions, where \methodname{} applies visual enhancement operations such as zoom, crop, and rotate to improve readability and increase recognition accuracy in these regions.

\paragraph{Document Question Answering.}
As shown in Table~\ref{tab:main_results}(b), \methodname{} consistently achieves superior performance across different backbone models on MMLongBench-Doc. With Gemini-2.5-Pro, \methodname{} reaches the best overall score of 61.5, outperforming the Gemini-2.5-Pro baseline by 3.4 points and all specialized pipelines. These improvements stem from two key designs. First, compared with general VLMs, \methodname{} leverages document processing tools to actively locate and inspect relevant content. Second, \methodname{} iteratively explores the document and dynamically assesses whether the evidence accumulated in the document state is sufficient to answer the question, thereby efficiently guiding subsequent exploration based on the current document state.

\paragraph{Key Information Extraction.}
We report the KIE results in Table~\ref{tab:main_results}(c). \methodname{} achieves competitive performance across all backbone models, reaching the best overall score of 86.9 and outperforming the specialized Qianfan-OCR by 2.9 points.
These improvements are attributed to the KIE document skill, which guides the agent to extract target field-value pairs through visual and OCR-based extraction, together with image-grounded verification to resolve inconsistent pairs.

\subsection{Ablation Studies}
\label{sec:experiments_analysis}

\begin{figure}[t!]
\centering
\includegraphics[width=\columnwidth]{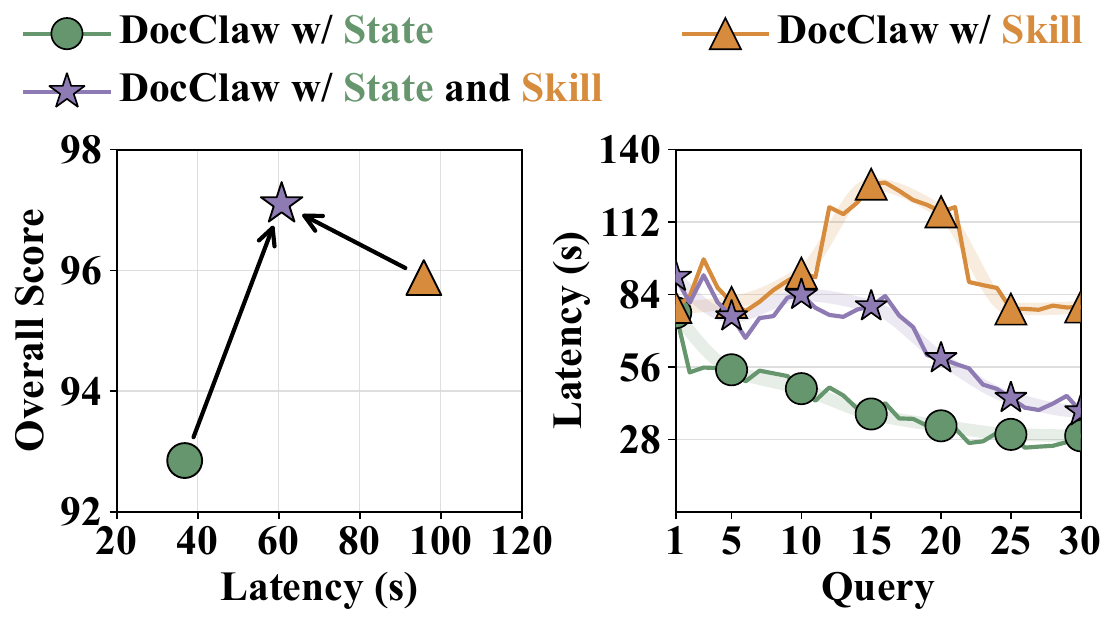}
\caption{
Ablation of document state and document skills.
Left: overall score and average latency when using document state, document skills, or both.
Right: latency across 30 consecutive queries on the same document.
}
\label{fig:ablation_latency}
\end{figure}

\paragraph{Effect of Document State and Skills.}

We study the effects of document state and document skills on a manually constructed set of 30 randomly shuffled queries over the same document, using Gemini-2.5-Flash as the backbone. The set comprises 10 queries each for OCR, DocQA, and KIE.
As shown in Figure~\ref{fig:ablation_latency}, adding document skills to the state-enabled variant improves the overall score from 92.85 to 97.10, demonstrating the benefit of task-oriented guidance, at the cost of additional processing latency.
In contrast, adding document state to the skill-enabled variant reduces the latency from 95.81s to 60.70s.
The right panel shows that the latency of state-enabled variants decreases as more queries are processed, as accumulated document knowledge can be reused by subsequent queries to avoid repeated processing.

\begin{figure}[t]
    \centering
    \includegraphics[width=0.975\columnwidth]{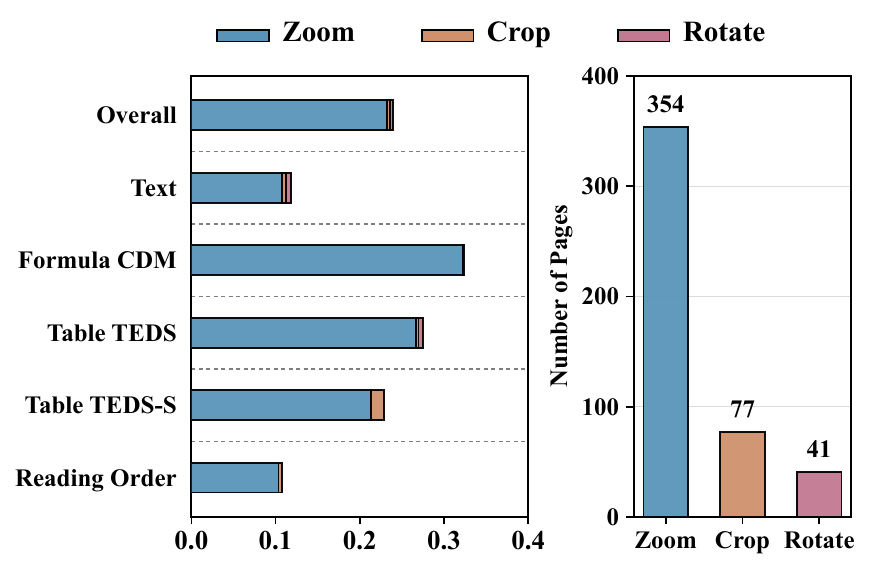}
    \caption{
Ablation of OCR refinement.
Left: performance improvement attributed to zoom, crop, and rotate operations.
Right: number of pages refined by each operation.
    }
    \label{fig:ocr_refinement_ablation}
\end{figure}

\paragraph{Effect of OCR Refinement.}
We analyze the contribution of refinement operations invoked during optical character recognition, including zoom, crop, and rotate.
For pages involving multiple operations, we evenly distribute the performance improvement among all applied operations.
As shown in Figure~\ref{fig:ocr_refinement_ablation}, refinement is triggered on 433 pages, with zoom applied most frequently to 354 pages, compared with 77 for crop and 41 for rotate.
Zoom contributes most of the overall improvement, with an attributed gain of 0.23. It is particularly effective for formula recognition, where zooming improves the visibility of small mathematical symbols.
Crop and rotate operations benefit table parsing by correcting region boundaries and orientations for more reliable recognition results.

\paragraph{Effect of KIE Verification.}

\begin{figure}[t!]
    \centering
    \includegraphics[width=0.98\columnwidth]{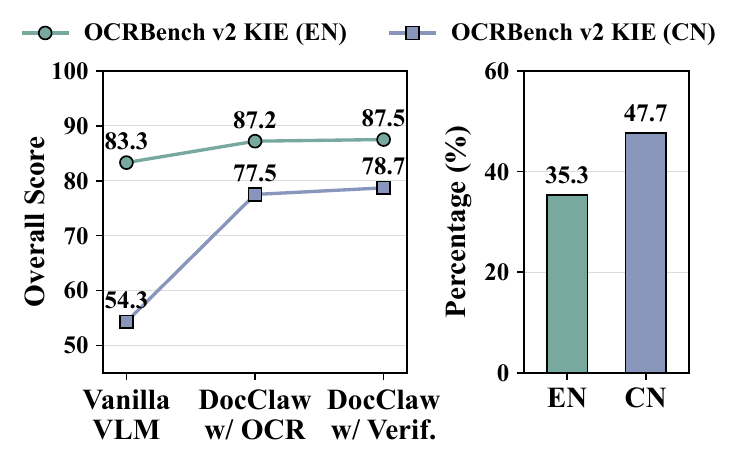}
    \caption{
Ablation of OCR and verification for KIE.
Left: KIE performance of the vanilla VLM and DocClaw with OCR and verification operations.
Right: percentage of vanilla VLM extraction errors corrected by the verification operation.
    }
    \label{fig:ablation_kie}
\end{figure}

We also study the effect of OCR and verification operations in the KIE task on OCRBench v2, using Gemini-2.5-Pro as the vanilla VLM baseline.
As shown in Figure~\ref{fig:ablation_kie}, incorporating OCR improves the KIE score from 83.3 to 87.2 on the English subset and from 54.3 to 77.5 on the Chinese subset, while verification further improves the scores to 87.5 and 78.7, respectively.
Notably, verification corrects 35.3\% and 47.7\% of the extraction errors made by the vanilla VLM on the English and Chinese subsets.
These results show that the verification operation can effectively correct erroneous field predictions.

%% file: sections/conclusion.tex
\section{Conclusion}
\label{sec:conclusion}

We introduced \methodname{}, a unified agentic system that formulates intelligent document processing as interactions between an agent and a document.
\methodname{} iteratively invokes processing tools under the guidance of document skills while maintaining a structured document state that organizes reusable document knowledge and task-specific interaction context.
Extensive experiments demonstrate competitive performance against general-purpose VLMs and task-specific methods.
Further ablation studies validate the effectiveness of document state and skills while demonstrating the contribution of task-specific refinement and verification operations.
